# Applying Incremental Learning in Binary-Addition-Tree Algorithm for Dynamic Binary-State Network Reliability

Wei-Chang Yeh

Department of Industrial Engineering and Engineering Management
National Tsing Hua University
P.O. Box 24-60, Hsinchu, Taiwan 300, R.O.C.
yeh@ieee.org, (+886)35742443

*Abstract* — This paper presents a novel approach to enhance the Binary-Addition-Tree algorithm (BAT) by integrating incremental learning techniques. BAT, known for its simplicity in development, implementation, and application, is a powerful implicit enumeration method for solving network reliability and optimization problems. However, it traditionally struggles with dynamic and large-scale networks due to its static nature. By introducing incremental learning, we enable the BAT to adapt and improve its performance iteratively as it encounters new data or network changes. This integration allows for more efficient computation, reduced redundancy without searching minimal paths and cuts, and improves overall performance in dynamic environments. Experimental results demonstrate the effectiveness of the proposed method, showing significant improvements in both computational efficiency and solution quality compared to the traditional BAT and indirect algorithms, such as MP-based algorithms and MC-based algorithms.

## 1. INTRODUCTION

In today's rapidly evolving technological landscape, ensuring network reliability has become a critical concern across diverse sectors such as telecommunications [1-2], power grids [3-5], transportation systems [6-8], and emerging fields like the Internet of Things (IoT) [9-10] and smart cities [11]. Network reliability, defined as the probability that a network operates at or above a specified performance level, is fundamental to evaluating network performance and guaranteeing uninterrupted services [12-13].

As our reliance on interconnected systems grows [14-15], so does the need for advanced methods to analyze and enhance network reliability. These techniques not only safeguard against unforeseen issues but also optimize resource allocation [12, 16-17], improve system efficiency, and foster technological innovation. However, the challenges in this field are numerous, encompassing increasing network complexity and scale, as well as the dynamic nature of threats and failure modes.

To address these challenges, researchers have developed various algorithms and methods, which can be broadly categorized into indirect and direct approaches:

- Indirect methods involve identifying minimal paths (MPs) [7, 15, 18] or minimal cuts (MCs) [19-20] and applying sum-of-disjoint products [21] or inclusion-exclusion techniques [22] to calculate exact reliability for binary-state networks. However, these methods face two NP-hard problems: finding MPs/MCs and implementing sum-of-disjoint products or inclusion-exclusion techniques.

- Direct methods, such as the Binary-Addition-Tree algorithm (BAT), have been developed to address these computational challenges [23-25]. BAT offers advantages in programming simplicity, adaptability, and efficiency compared to traditional search methods.

Despite these advancements, both indirect and direct methods, including BAT, are primarily designed for static networks [28, 18, 19, 41, 42]. This static nature presents significant limitations when applying these algorithms to dynamic networks, which frequently undergo changes due to component failures, repairs, upgrades, or expansions [28, 18, 19, 41, 42]. In dynamic environments, static optimization approaches are inefficient, requiring complete reinitialization and reoptimization whenever changes occur [30, 31, 32].

To overcome these limitations, this paper proposes integrating incremental learning into the BAT. Incremental learning, a machine learning paradigm that continuously updates models as new data becomes available, enables adaptation to change without complete retraining. This integration aims to enhance BAT's performance in dynamic networks by allowing it to dynamically update pheromone trails—probabilistic markers that guide the search process—thereby maintaining efficiency and accuracy even as networks evolve.

The proposed incremental learning-enabled BAT (IL-BAT) addresses the inefficiencies of traditional methods in dynamic environments, improving scalability and robustness in network reliability analysis. This approach is particularly valuable for large-scale networks and real-time applications, where timely and reliable solutions are crucial. By integrating incremental learning into BAT, this work represents a significant advancement in ensuring the reliability of critical infrastructures under dynamic and unpredictable conditions.

The paper is structured as follows: Section 2 outlines the necessary acronyms, notations, nomenclature, and assumptions. Section 3 provides background information on BAT, LSA/PLSA, and incremental learning. Section 4 discusses the innovative aspects of the proposed IL-BAT, while Section 5 details the methodology for integrating incremental learning with BAT, covering both the theoretical foundations and practical applications, along with comparative performance analysis. Section 6 concludes with a discussion of the findings and suggestions for future research directions.

Overall, this work contributes to the ongoing development of more adaptive and efficient optimization algorithms, paving the way for enhanced network reliability in an increasingly interconnected world.

## 2. ACRONYMS, NOTATIONS, NOMENCLATURE, AND ASSUMPTIONS

This section provides an overview of the key acronyms, notations, nomenclature, and assumptions relevant to the proposed IL-BAT.

### 2.1 Acronyms

BAT：Binary-Addition-Tree Algorithm

IL-BAT：Incremental learning-enabled BAT

LSA：Layered search algorithm

MP :Minimal path

MC :Minimal cut

### 2.2 Notations

$|\bullet|$：number of elements in $\bullet$

$\Pr(\bullet)$：probability of $\bullet$

$a_i$：arc $i$

$p_i$：The probability $\Pr(a_i)$ that $a_i$ is in a working state.

$q_i$：The probability $1-\Pr(a_i)$ that $a_i$ is in a failed state, where $p_i + q_i = 1$.

$V$: Node set $V = \{1, 2, \ldots, n\}$

$E$: Arc set $E = \{a_1, a_2, \ldots, a_m\}$

$n$: The number of nodes, $|V|$.

$m$: The number of arcs, $|E|$.

$X$: A binary-state vector.

$X(a_i)$: The value of arc $a_i$ in the vector $X$, for $i = 1, 2, \ldots, m$.

$B(\bullet)$: The vector set $B(\bullet) = \{\, X_i \mid X_i$ is the $i$-th binary-state vector derived from the BAT for the set $\bullet$, where $i = 1, 2, \ldots, 2^{|\bullet|}\}$.

$\mathbf{D}$: The binary-state distribution of arcs, listing the probabilities $p_i$ for all arcs $a_i$.

$G(V, E)$: The undirected original graph defined by its sets of nodes $V$ and edges $E$. For example, Figure 1 illustrates such a graph before any incremental learning process.

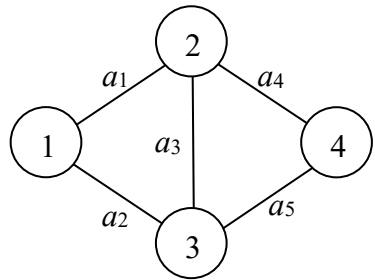

**Figure 1.** An example graph.

$G(V, E, \mathbf{D})$: An original graph $G(V, E)$ with an associated binary-state distribution $\mathbf{D}$. For example, Figure 1 represents a binary-state network when the distribution $\mathbf{D}$ is specified.

$E(X)$: The subset of arcs induced by the binary-state $X$, defined as $E(X) = \{a \in E \mid X(a) = 1\}$.

$V(X)$: The subset of nodes induced by the binary-state $X$, defined as $V(X) = \{v \in V \mid X(a) = 1$ and $a$ is adjacent to $v\}$.

$G(X)$: The subgraph corresponding to $X$, defined as $G(X) = G(V, E(X), \mathbf{D})$.

$P_i$: The arc or path added during the $i$-th incremental learning process.

$\Lambda$: The number of incremental learning processes.

$V(P)$: The subset of nodes induced by the arc or path $P$, defined as $V(P) = \{v \in V \mid a \in P$ and $a$ is adjacent to $v\}$.

$\otimes$: The convolution product defined as $X \otimes Y = (X(\alpha_1), X(\alpha_2), \ldots, X(\alpha_v), Y(\beta_1), Y(\beta_2), \ldots, Y(\beta_\mu))$, where $X = (X(\alpha_1), X(\alpha_2), \ldots, X(\alpha_v))$ and $Y = (Y(\beta_1), Y(\beta_2), \ldots, Y(\beta_\mu))$.

$G(X) \oplus P$: The incremental network after adding the incremental learning process $P$ to the subgraph $G(X)$; this is also denoted as $G(X \otimes P)$.

$R_i$: The reliability after the $i$-th incremental learning process, where $R_0$ is the reliability of the original network $G(V, E, \mathbf{D})$.

$L_i$: The $i$th layer obtained from the LSA/PLSA, where $i = 1, 2, \ldots, m$.

$S(X)$: The set of nodes $S(X) = \{v \in V(X) |$ nodes 1 and $v$ is connected in the subnetwork $G(X)\}$.

$T(X)$: The set of nodes $T(X) = \{ v \in V(X) |$ nodes $v$ and $n$ is connected in the subnetwork $G(X)\}$.

$M(X)$: $M(X) = V - [S(X) \cup T(X)]$.

## 2.3 Nomenclature

Binary-state network: A network in which each arc can exist in only two states: 0 (failed) or 1 (operational).

Reliability: The probability that a network is operating successfully.

Feasible vector: A vector $X$ is considered feasible if nodes 1 (source) and $n$ (sink) are connected in $G(X)$.

Infeasible vector: A vector $X$ is considered infeasible if there is no existing path from 1 (source) to $n$ (sink) in $G(X)$.

Initial/Original network: The original network, represented as a binary-state network $G(V, E, \mathbf{D})$, where $V$ is the set of nodes, $E$ is the set of arcs, and $\mathbf{D}$ denotes the binary-state distribution of the arcs.

Incremental learning process: As changes occur in the network, such as the addition of new paths, arcs, or nodes, these modifications are incorporated into the existing network incrementally. Each change, or "incremental learning process," represents an update to the network's structure.

Incremental Network: After each incremental learning process, a new version of the network is formed, referred to as the incremental network. For example, if a new path $P_1$ is added to the original network $G(X)$, the resulting network, denoted $G(X) \oplus P_1$, is an incremental network. This process can continue with further increments, such as adding another path $P_2$, resulting in a new network $G(X) \oplus P_1 \oplus P_2$.

## 2.4 Assumptions

The following assumptions form the basis of our analysis of binary-state networks:

1. Simple graph structure: The network is modeled as a simple graph without parallel arcs (multiple edges between the same pair of nodes) or self-loops (edges that connect a node to itself).

2. Perfect node reliability: All nodes in the network are assumed to be fully reliable and always connected. This assumption allows the analysis to focus solely on arc failures without considering node failures as potential causes of network degradation.

3. Arc state independence: The states of individual arcs are statistically independent, meaning the state of one arc does not affect or depend on the state of any other arc in the network.

## 3. REVIEW OF BAT, PLSA, AND INCREMENTAL LEARNING

The proposed IL-BAT is designed to efficiently calculate reliability in dynamic binary-state networks by handling a series of incremental learning processes. For the original network, it uses the BAT method [24] to generate feasible vectors for reliability calculation and adapts the PLSA to effectively group arcs. It then employs sub-BAT and the convolution product to extend disconnected vectors to the incremental networks. This section provides a brief overview of these foundational components, establishing the necessary context for the innovations introduced by the IL-BAT.

## 3.1 BAT

Yeh's BAT is a highly efficient method for generating all possible binary-state vectors, where each coordinate can take on a value of either 0 or 1 [24]. The algorithm is based on two fundamental rules that streamline the generation process [23-25].

The first rule involves identifying the first coordinate in the vector that has a value of zero, denoted as $x_i$ [24]. Once this zero coordinate is found, the algorithm sets all preceding coordinates $x_j$ for $j < i$ to zero. This approach ensures that the algorithm systematically explores all possible combinations of binary states by resetting parts of the vector while maintaining a logical sequence in its generation process.

The second rule dictates that the process should terminate when the last coordinate in the vector is reached, indicating that all possible binary-state vectors have been generated [24]. This rule is crucial as it provides a clear stopping point, ensuring that the algorithm efficiently covers the entire solution space without redundancy.

Together, these two straightforward yet powerful rules form the foundation of the BAT [24]. By adhering to them, the algorithm can generate binary-state vectors in a methodical and efficient manner, making it highly effective for applications that require exhaustive exploration of binary states.

The pseudocode of the BAT encapsulates these principles, providing a clear framework for implementation. For those interested in exploring the algorithm further, the full source code is available for download, as referenced in [26]. This code offers a practical example of how the BAT can be applied to various computational problems, particularly those involving network reliability [23, 25, 27], resilience analysis [28-31], and other areas where binary-state vectors play a critical role [32-33].

**Algorithm: BAT**

**Input:**      $m$ (the number of coordinates).

**Output:**   All vectors.

**STEP 0.**   Initialize the vector $X$ as an $m$-tuple zero.

**STEP 1.**   Locate the zero coordinate, denoted as $a_k$, let $x_k = x_k + 1$ and $x_j = 0$ for $i = 1, 2, \ldots, (k-1)$.

**STEP 2.**   If no such coordinate is found, terminate the process as all vectors have been obtained. If a coordinate was found, return to STEP 1.

The algorithm begins by initializing $X = (x_1, x_2, \ldots, x_m)$ to zero and setting $i$ to 1. It then enters a main loop where it checks whether $x_i$ is zero. If it is, the algorithm sets $x_i$ to 1 and resets $i$ to 1, following the first rule mentioned earlier. The loop continues until $i$ equals $m$, indicating the last coordinate has been reached. At this point, according to the second rule, the algorithm terminates. If $i$ has not yet reached $m$, the algorithm sets $x_i$ back to zero, increments $i$ by 1, and the loop resumes.

The elegance of BAT lies in its four-step iterative process, which repeatedly updates a single $m$-tuple binary-state vector $X$. This approach offers several advantages: it is straightforward to implement, efficient in execution, economical in terms of memory usage, and flexible enough to be adapted to various applications.

Due to its versatility, BAT has been widely adopted in different fields. Researchers have applied it to calculate network reliability [23, 25, 27], assess resilience [28-31], estimate the probability of wildfire spread [32], and analyze the propagation probability of computer viruses [34]. Its ability to efficiently explore all possible binary-state vectors makes it a powerful tool in these domains [23, 25, 27-32, 34].

To illustrate how the algorithm operates, consider the generation of all 5-tuple binary-state vectors $X$ using the BAT pseudocode. Table 1 presents these vectors, with $X$ representing the vector at each iteration. Although the subscript $i$ is included for clarity, it is not essential for the algorithm's functioning. The final column of the table indicates whether $X_i$ is feasible. This example demonstrates the practicality of BAT in generating binary-state vectors and identifying feasible vectors within a network.

**Table 1.** 5-tuple vectors via BAT.

| $i$ | $x_1$ | $x_2$ | $x_3$ | $x_4$ | $x_5$ | $i$ | $x_1$ | $x_2$ | $x_3$ | $x_4$ | $x_5$ |
|---|---|---|---|---|---|---|---|---|---|---|---|
| 1 | 0 | 0 | 0 | 0 | 0 | 17 | 0 | 0 | 0 | 0 | 1 |
| 2 | 1 | 0 | 0 | 0 | 0 | 18 | 1 | 0 | 0 | 0 | 1 |
| 3 | 0 | 1 | 0 | 0 | 0 | 19 | 0 | 1 | 0 | 0 | 1 |
| 4 | 1 | 1 | 0 | 0 | 0 | 20 | 1 | 1 | 0 | 0 | 1 |
| 5 | 0 | 0 | 1 | 0 | 0 | 21 | 0 | 0 | 1 | 0 | 1 |
| 6 | 1 | 0 | 1 | 0 | 0 | 22 | 1 | 0 | 1 | 0 | 1 |
| 7 | 0 | 1 | 1 | 0 | 0 | 23 | 0 | 1 | 1 | 0 | 1 |
| 8 | 1 | 1 | 1 | 0 | 0 | 24 | 1 | 1 | 1 | 0 | 1 |
| 9 | 0 | 0 | 0 | 1 | 0 | 25 | 0 | 0 | 0 | 1 | 1 |
| 10 | 1 | 0 | 0 | 1 | 0 | 26 | 1 | 0 | 0 | 1 | 1 |
| 11 | 0 | 1 | 0 | 1 | 0 | 27 | 0 | 1 | 0 | 1 | 1 |
| 12 | 1 | 1 | 0 | 1 | 0 | 28 | 1 | 1 | 0 | 1 | 1 |
| 13 | 0 | 0 | 1 | 1 | 0 | 29 | 0 | 0 | 1 | 1 | 1 |
| 14 | 1 | 0 | 1 | 1 | 0 | 30 | 1 | 0 | 1 | 1 | 1 |
| 15 | 0 | 1 | 1 | 1 | 0 | 31 | 0 | 1 | 1 | 1 | 1 |
| 16 | 1 | 1 | 1 | 1 | 0 | 32 | 1 | 1 | 1 | 1 | 1 |

## 3.3 LSA and PLSA

Network reliability, defined as the probability of successfully connecting the source node and the sink node under specific conditions—such as resource limitations, time constraints, capacity constraints, risk thresholds, reliability thresholds, operational constraints, energy consumption limits, data transmission limits, failure tolerance levels, and security requirements—is a critical concept across various fields [6, 21-24]. Verifying this connectivity is essential for solving network reliability problems.

The Layered Search Algorithm (LSA) has emerged as a powerful tool in this area. Initially designed to identify all $d$-MPs (minimal paths) in acyclic multi-state flow networks [35], LSA has since been adapted for multiple applications, including verifying connectivity in binary-state graphs $G(V, E)$ [36] and effectively grouping arcs [37]. LSA operates with a time complexity of $O(|V|)$ and is based on the Breadth-First Search method. It constructs layers where each node in the current layer is connected to at least one node in the preceding layer.

Building on LSA, the Path-based Layer-Search Algorithm (PLSA) was developed as a modification to enhance efficiency, particularly in checking the connectivity of nodes 1 and $n$ within BATs [24]. Due to its effectiveness, PLSA has been incorporated into the proposed algorithm for connection verification.

PLSA functions by constructing disjoint node subsets $L_1, L_2, ..., L_\lambda$ sequentially [24]. This process continues until either no more layers can be formed or node $n$ is included in the most recently discovered layer. Each layer $L_i$ is defined as $\{v \mid a \in E$, for $a$ is adjacent to nodes $u \in L_{(i-1)}$ and $v \notin \cup_{k=1}^{(i-1)} L_k\}$. The algorithm concludes that nodes 1 and $n$ are connected if $n \in L_\lambda$; otherwise, no connection exists.

The pseudocode of the PLSA is listed in the following:

**Algorithm: PLSA**

**Input:**     A graph $G(V, E)$ and vector $X$.

**Output:**   Determine whether $X$ is connected, i.e., if there is a directed path from node 1 to node $n$ in $G(X)$.

**STEP 0.**   Initialize $i = 2$ and the first layer $L_1 = \{1\}$.

**STEP 1.**   Define $L_i = \{v \mid a \in E$, for $a$ is adjacent to nodes $u \in L_{(i-1)}$ and $v \notin \cup_{k=1}^{(i-1)} L_k\}$.

**STEP 2.**   If $n \in L_i$, stop; $X$ is connected.

**STEP 3.** If $L_i = \varnothing$, $X$ is disconnected.

**STEP 4.** Increment $i$ by 1 and return to STEP 1.

Consider the network depicted in Figure 1, where the arc-state vector $X$ is represented as (1, 1, 1, 1, 1). The layered structure of $G(X)$ is defined as follows: $L_1 = \{1\}$, $L_2 = \{2, 3\}$, and $L_3 = \{4\}$. This configuration demonstrates the feasibility of $X$. Moreover, the network exhibits full connectivity, with all nodes $\{1, 2, 3, 4\}$ forming a connected component.

## 3.3 Incremental Learning

Incremental learning is a machine learning approach that continuously updates models as new data arrives, rather than retraining from scratch. This method is ideal for scenarios where data is generated over time or where systems need dynamic adaptation. Incremental learning is applied in areas like data stream mining, adaptive control systems, personalized recommendations, and predictive maintenance, demonstrating its flexibility and effectiveness across domains.

The efficacy of incremental learning is underpinned by several key attributes:

1. Computational Efficiency: It demonstrates reduced computational and memory requirements compared to batch learning, facilitating real-time application deployment.

2. Adaptive Capability: The methodology exhibits rapid adjustment to evolving data patterns or system conditions, thereby maintaining model accuracy.

3. Scalability: Large-scale datasets are effectively managed through incremental model updates, mitigating the need for extensive computational resources.

4. Perpetual Learning: Models can continuously evolve and improve, a critical feature in dynamic data environments.

This study represents a significant milestone in network reliability by applying incremental learning techniques to this domain for the first time. It demonstrates how incremental updates, such as adding new paths — $P_1 = \{a_6, a_7\}$, introduced in Figure 1 and resulting in the modified network in Figure 2(a), and $P_2 = \{a_8\}$, subsequently added in Figure 2(a) to produce the further updated configuration in Figure 2(b)—can dynamically adapt network structures.

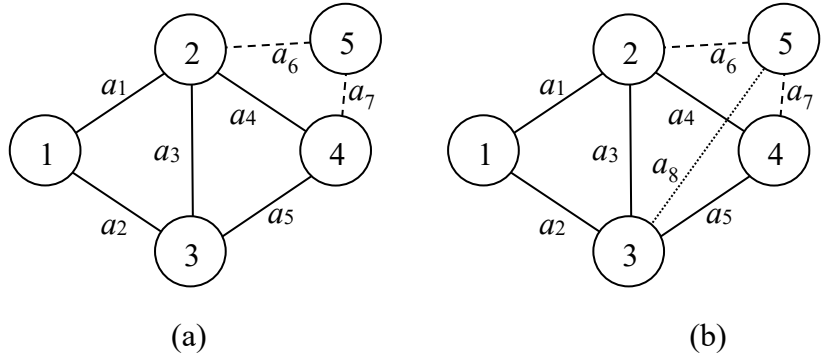

(a)                                    (b)

**Figure 2**. Networks representing $G(V, E, \mathbf{D}) \oplus P_1$ and $G(V, E, \mathbf{D}) \oplus P_1 \oplus P_2$.

Integrating incremental learning with network reliability analysis provides an innovative approach to addressing the dynamic complexities of modern networks. By allowing real-time adaptations to network changes without requiring complete model retraining, this research establishes new paradigms for enhancing network resilience and performance. The practical implications are extensive, potentially transforming network reliability assessment and management across sectors such as telecommunications, transportation, and critical infrastructure systems.

## 4. PROPOSED INNOVATIONS

The proposed IL-BAT introduces critical innovations to enhance efficiency over the traditional BAT algorithm. A key improvement is the method of splitting the node set for each vector, which is crucial for the algorithm's performance. Unlike the traditional BAT, which identifies all possible binary-state vectors including both feasible and infeasible solutions, IL-BAT improves efficiency by excluding a larger number of feasible vectors and implementing a quick connect check. These major innovations, aimed at achieving the dual goals of reducing the solution space and accelerating connectivity verification, are central to IL-BAT's enhanced performance and are discussed in detail in this section.

### 4.1 The sub-BAT for $P$, $\oplus$, and $\otimes$

Let $X$ be an infeasible vector in $G(V, E, \mathbf{D})$, let $P$ represent an incremental learning process. Define $E(X)$ as the set of arcs with a coordinate of one in vector $X$, and $V(P)$ as the set of nodes involved in process $P$. For example, in Figure 1, $E(1, 1, 1, 0, 0) = \{a_1, a_2, a_3\}$. In Figure 2(a) and Figure 2(b), $P_1 = \{a_6, a_7\}$ with $V(P_1) = \{2, 4, 5\}$, and $P_1 = \{a_8\}$ with $V(P_2) = \{3, 5\}$ respectively.

The convolution product proposed in [38] is adapted here to create a new vector $X^* = X \otimes Y$, derived from two vectors $X$ and $Y$, such that there are no common coordinates between arc vectors $X$ and $Y$. Specifically, for any arc $a$, $X^*(a) = X(a)$ if $a \in X$, and $X^*(a) = Y(a)$ if $a \in Y$.

$G(X)$ denotes the subnetwork induced by arc vector $X$, defined as $G(X) = G(V, E(X), \mathbf{D})$. The notation $G(X) \oplus Y$ represents adding an arc subset $E(Y) \subseteq P$ to $G(X)$, resulting in the new network $G(X) \oplus Y = G(X \otimes Y)$.

To accurately determine network reliability, each feasible vector must be identified, even after each incremental learning process. For every incremental learning process $P$ applied to $G(X)$, we need to verify whether any subset $Y \subseteq P$ already connects nodes 1 and $n$ in the network $G(X) \oplus Y = G(X \otimes Y)$. Therefore, all combinations of each incremental learning process $P$ must be considered before extending a disconnected vector $X$ to new vector $X \otimes Y$.

The BAT algorithm is implemented to achieve this objective, and this specific implementation is referred to as the "sub-BAT" to indicate that it is tailored for process $P$. For example, in Figure 2(a), with $P_1 = \{a_6, a_7\}$, we derive the combinations $Y = (x_6, x_7) = (0, 0), (1, 0), (0, 1)$, and $(1, 1)$ from the BAT.

## 4.2 Determining the Initial S(X), T(X) and M(X) using the PLSA

A crucial innovation in the IL-BAT is the concept to split the node set $V$. For any binary-state vector $X$, the algorithm defines three distinct subsets of nodes:

- $S(X)$: The set of nodes connected to the source node 1 in the subnetwork $G(X)$.

- $T(X)$: The set of nodes connected to the sink node $n$ in the subnetwork $G(X)$.

- $M(X)$: The set of nodes that are not included in either $S(X_i)$ or $T(X_i)$, defined as $M(X_i) = V - [S(X_i) \cup T(X_i)]$.

To perform this node-splitting, the algorithm utilizes a new application of PLSA. This process is systematically applied to all possible vectors $X_i$, where $i$ ranges from 1 to $2^m$ (with $m$ being the number of arcs in the original network before any incremental learning process). Figure 1 and Table 2 illustrate this concept with a sample network and show the resulting subsets of nodes for each vector.

This technique of splitting nodes is essential to the improved efficiency of IL-BAT, as it allows for a more refined analysis of network connectivity and significantly reduces the number of vectors that need a full evaluation.

For example, in the network depicted in Figure 1, Table 2 lists all the vectors $X_i$ along with their corresponding subsets $S(X_i)$, $M(X_i)$, and $T(X_i)$.

**Table 2.** The sets $S(X_i)$, $M(X_i)$, and $T(X_i)$ for each $X_i$ in Figure 1.

| $i$ | $X_i=(x_1, x_2, x_3, x_4, x_5)$ | $S(X_i)$ | $M(X_i)$ | $T(X_i)$ | Feasible? |
|---|---|---|---|---|---|
| 1 | (0, 0, 0, 0, 0) | {1} | {2, 3} | {4} | |
| 2 | (1, 0, 0, 0, 0) | {1, 2} | {3} | {4} | |
| 3 | (0, 1, 0, 0, 0) | {1, 3} | {2} | {4} | |
| 4 | (1, 1, 0, 0, 0) | {1, 2, 3} | $\varnothing$ | {4} | |
| 5 | (0, 0, 1, 0, 0) | {1} | {2, 3} | {4} | |
| 6 | (1, 0, 1, 0, 0) | {1, 2, 3} | $\varnothing$ | {4} | |
| 7 | (0, 1, 1, 0, 0) | {1, 2, 3} | $\varnothing$ | {4} | |
| 8 | (1, 1, 1, 0, 0) | {1, 2, 3} | $\varnothing$ | {4} | |
| 9 | (0, 0, 0, 1, 0) | {1} | {3} | {2, 4} | |
| 10 | (1, 0, 0, 1, 0) | {1, 2, 4} | $\varnothing$ | {1, 2, 4} | |
| 11 | (0, 1, 0, 1, 0) | {1, 3} | $\varnothing$ | {2, 4} | |
| 12 | (1, 1, 0, 1, 0) | {1, 2, 3, 4} | $\varnothing$ | {1, 2, 3, 4} | Y |
| 13 | (0, 0, 1, 1, 0) | {1} | $\varnothing$ | {2, 3, 4} | |
| 14 | (1, 0, 1, 1, 0) | {1, 2, 3, 4} | $\varnothing$ | {1, 2, 3, 4} | Y |
| 15 | (0, 1, 1, 1, 0) | {1, 2, 3, 4} | $\varnothing$ | {1, 2, 3, 4} | Y |
| 16 | (1, 1, 1, 1, 0) | {1, 2, 3, 4} | $\varnothing$ | {1, 2, 3, 4} | Y |
| 17 | (0, 0, 0, 0, 1) | {1} | $\varnothing$ | {3, 4} | |
| 18 | (1, 0, 0, 0, 1) | {1, 2} | $\varnothing$ | {3, 4} | |
| 19 | (0, 1, 0, 0, 1) | {1, 3, 4} | {2} | {1, 3, 4} | Y |
| 20 | (1, 1, 0, 0, 1) | {1, 2, 3, 4} | $\varnothing$ | {1, 2, 3, 4} | Y |
| 21 | (0, 0, 1, 0, 1) | {1} | $\varnothing$ | {2, 3, 4} | |
| 22 | (1, 0, 1, 0, 1) | {1, 2, 3, 4} | $\varnothing$ | {1, 2, 3, 4} | Y |
| 23 | (0, 1, 1, 0, 1) | {1, 2, 3, 4} | $\varnothing$ | {1, 2, 3, 4} | Y |
| 24 | (1, 1, 1, 0, 1) | {1, 2, 3, 4} | $\varnothing$ | {1, 2, 3, 4} | Y |
| 25 | (0, 0, 0, 1, 1) | {1} | $\varnothing$ | {2, 3, 4} | |
| 26 | (1, 0, 0, 1, 1) | {1, 2, 3, 4} | $\varnothing$ | {1, 2, 3, 4} | Y |
| 27 | (0, 1, 0, 1, 1) | {1, 2, 3, 4} | $\varnothing$ | {1, 2, 3, 4} | Y |
| 28 | (1, 1, 0, 1, 1) | {1, 2, 3, 4} | $\varnothing$ | {1, 2, 3, 4} | Y |
| 29 | (0, 0, 1, 1, 1) | {1} | $\varnothing$ | {2, 3, 4} | |
| 30 | (1, 0, 1, 1, 1) | {1, 2, 3, 4} | $\varnothing$ | {1, 2, 3, 4} | Y |
| 31 | (0, 1, 1, 1, 1) | {1, 2, 3, 4} | $\varnothing$ | {1, 2, 3, 4} | Y |
| 32 | (1, 1, 1, 1, 1) | {1, 2, 3, 4} | $\varnothing$ | {1, 2, 3, 4} | Y |

**4.3 Verify the Connectivity of $G(X)$ by Testing if $S(X) = T(X)$**

The IL-BAT algorithm enhances network reliability analysis through efficient connectivity verification and incremental learning. Network reliability is calculated by summing the probabilities of all connected vectors derived from BAT vectors. While the PLSA can verify connectivity after adding a new path $P$ to a disconnected graph, IL-BAT introduces a more efficient method using $S(X)$ and $T(X)$ sets.

IL-BAT exploits a fundamental property of network connectivity: in a connected network, any node linked to the source (node 1) is also connected to the sink (node n), and vice versa. This implies that for any feasible vector $X$ where nodes 1 and $n$ are connected in $G(X)$, the sets $S(X)$ and $T(X)$) are identical. Table 2 illustrates this property, showing that $S(X_i) = T(X_i)$ for $i$ = 12, 14, 15, 16, 20, 22, 23, 24, 26, 27, 28, 30, 31, 32.

The time complexity for checking the equality of two sets, $S(X)$ and $T(X)$, is determined by the size of the smaller set, i.e., $O(\min\{|S(X)|, |T(X)|\})$. Since both $|S(X)|$ and $|T(X)|$ represent subsets of the node set $V$, their sizes cannot exceed $|V|$. Therefore, checking whether $|S(X)| = |T(X)|$ is more efficient than the $O(|V|)$ time required by PLSA to determine if $G(X)$ is connected.

This property remains consistent during incremental learning. When a new path $P$ is added to $G(X)$, creating $G(X) \oplus P$, if $G(X)$ was initially connected, it remains connected in $G(X) \oplus P$. Consequently, $S(X \otimes P)$ and $T(X \otimes P)$ remain equal for each feasible $X \oplus P$ in the expanded graph $G(X \otimes P)$. Figures 2(a) and 2(b) demonstrate this concept, showing that connectivity is maintained after incremental updates to the initially connected graph in Figure 1.

To optimize performance and memory usage, the IL-BAT employs a streamlined approach. It verifies the feasibility of each vector, adds its probability to the overall reliability $R$ if feasible, and then discards the vector. This process is also exemplified in Table 2, where feasible vectors (those with $S(X_i) = T(X_i)$) have their probabilities calculated and added to $R$ before being immediately discarded. This approach significantly reduces memory requirements and accelerates the computation process by avoiding unnecessary storage of processed vectors.

### 4.4 Determining $S(X)$, $T(X)$, and $M(X)$ in the incremental network

The concept of network connectedness is central to the proposed IL-BAT. As outlined in Section 4.2, a network $G(X)$ is considered connected when $S(X)$, the reachable node subset of source node, is identical to $T(X)$, the reachable node subset of sink node. This equality specifically indicates that nodes 1 and $n$ are connected within $G(X)$. Conversely, when $S(X)$ and $T(X)$ differ, $G(X)$ is disconnected.

To illustrate this concept, consider the example in Figure 1. Here, we have $X_{11} = (0, 1, 0, 1, 0)$, which represents a disconnected network. In this case, $S(X_{11}) = \{1, 3\}$ and $T(X_{11}) = \{2, 4\}$, clearly demonstrating that $S(X_{11}) \neq T(X_{11})$. Additionally, $M(X_{11})$, the set of middle nodes, is empty.

Now, let's consider a scenario where we have a disconnected graph $G(X)$ and we want to apply an incremental learning process $P$. Section 4.2 introduces an important efficiency consideration: when determining the connectedness of $G(X) \oplus P$ (the graph resulting from adding $P$ to $G(X)$), it's more computationally efficient to simply check whether $S(X)$ equals $T(X)$, rather than employing the more complex PLSA.

This efficiency gain, however, presents a new challenge. We need to develop a method to efficiently update $S(X \otimes P)$, $T(X \otimes P)$, and $M(X \otimes P)$ after the addition of $P$ to $G(X)$. While it's possible to use LSA to redistribute all nodes in $X \otimes P$ across these three sets in $O(|X \otimes P|)$, a more efficient approach based on set operations is proposed.

The proposed method is elegantly simple: for each set $\phi$ in $\{S(X), T(X), M(X)\}$, if $\phi$ intersects with $V(P)$ (the set of nodes in the added component), we set $\phi \otimes P$ to the union of $\phi$ and $V(P)$. Mathematically, this is expressed as:

$$\text{Let } \phi \otimes P \text{ to } \phi \cup V(P) \text{ if } \phi \cap V(P) \neq \varnothing \text{ for all } \phi \text{ in } \{S(X), T(X), M(X)\}. \tag{1}$$

Assuming both $\phi$ and $V(P)$ are represented as hash sets (or similar data structures), the average time complexity for checking if any element of $V(P)$ is in $\phi$ is $O(\min(|\phi|, |V(P)|))$, as mentioned in Section 4.3. Additionally, performing the union operation $\phi \cup V(P)$ takes $O(|V(P)|)$. Therefore, the total time complexity to execute Eq. (1) is $O(\min(|S(X)|, |V(P)|) + \min(|T(X)|, |V(P)|) + \min(|M(X)|, |V(P)|) + 3|V(P)|) < O(|V| + |V(P)|)$. Thus, it is more efficient to compute $S(X \otimes P)$, $T(X \otimes P)$, and $M(X \otimes P)$ based on $S(X)$, $T(X)$, $M(X)$, respectively, rather than using the PLSA to find them.

To better understand this process, let's examine the example illustrated in Figure 2. In Figure 2(a), we start with the initial graph $G(X)$. Figure 2(a) then shows the result of adding $P = \{a_6, a_7\}$ to this graph. In this case, $V(P) = \{2, 4, 5\}$, representing the nodes in the added component.

Applying our update rule, we find that $T(X)$ intersects with $V(X)$, so $T(X \otimes P)$ is set to $T(X) \cup V(P) = \{2, 4, 5\} = \{2, 4, 5\}$. However, neither $S(X)$ nor $M(X)$ intersect with $V(X)$, so they remain unchanged. This example demonstrates how our efficient update method selectively modifies only the relevant sets, avoiding unnecessary computations.

This approach significantly streamlines the process of maintaining and updating the connectedness information as the graph evolves, providing a more efficient alternative to re-running the PLSA on the entire updated graph.

### 4.5 The Zero Vector

As mentioned in Section 4.1, all combinations need to be considered during each incremental learning process $P$. The zero vector is the first vector derived from the BAT for combinations across all sets. A specific procedure for handling the zero vector is described below.

Since $V(\mathbf{0}) = \varnothing$, the sets $S(X \otimes \mathbf{0})$, $T(X \otimes \mathbf{0})$, and $M(X \otimes \mathbf{0})$ are equal to $S(X)$, $T(X)$, and $M(X) \cup \{v \mid$ a node in $V(P)$ but not in the previous network$\}$, respectively. Additionally, if the network $G(X)$ is disconnected, then $G(X \otimes \mathbf{0})$ will also remain disconnected. Therefore, the zero vector does not need to be considered if it is known to be the final step in the incremental learning process.

For example, consider Figure 2(a), where the network is incrementally updated by adding $P_1 = \{a_6, a_7\}$ to Figure 1. For every infeasible vector $X$ derived in $G(V, E, \mathbf{D})$ (such as $X_{1,8} = (1, 1, 1, 0, 0)$), the following holds: $S(X \otimes \mathbf{0}) = S(X)$, $T(X \otimes \mathbf{0}) = T(X)$, and $M(X \otimes \mathbf{0}) = M(X) \cup \{v \mid v \in V(P_1)$ but $v \notin V\}$ For instance, for $X_{2,29} = X_{1,8} \otimes \mathbf{0} = (1, 1, 1, 0, 0, 0, 0)$, we have $S(X_{2,29}) = S(X_{1,8}) = \{1, 2, 3\}$, $T(X_{2,29}) = T(X_{1,8}) = \{4\}$, and $M(X_{2,29}) = M(X_{1,8}) \cup \{5\} = \{5\}$.

In another example, as shown in Figure 2(b), where the network is further incremented by adding $P_2 = \{a_8\}$ to Figure 2(a), all infeasible vectors derived in $G(V, E, \mathbf{D}) \oplus P_1$ (such as $X_{2,1} = (0, 0, 0, 0, 0, 0, 0)$) can ignore the convolution product with zero vector $(0)$ derived from $P_2$.

## 5. PROPOSED COMPLETE IL-BAT

Section 5 provides a comprehensive overview of the proposed IL-BAT. It presents the complete IL-BAT pseudocode and analyzes its time complexity, building on the foundations established in Section 4. To demonstrate practical application, the section includes a step-by-step implementation using the network illustrated in Figure 1. Additionally, it features a computational experiment comparing the number of obtained terms among the complete IL-BAT, traditional BAT, and other conventional algorithms. This thorough examination showcases the algorithm's structure, efficiency, and performance in dynamic network scenarios, offering readers a deep understanding of IL-BAT's capabilities in network reliability analysis.

## 5.1 Pseudocode and Time Complexity

The pseudocode for the complete IL-BAT integrates the traditional BAT, sub-BAT, and innovations from Section 4. It begins with initializing the network and performing an initial BAT calculation. When network changes occur, it identifies the type of change through the set operations and applies the appropriate incremental learning procedure.

For each incremental learning process, it updates the structure, calculates new vectors using sub-BAT, and merges them with existing ones. Probability modifications update arc probabilities and recalculate affected vectors without structural changes. The algorithm then updates the overall network reliability and repeats this process for subsequent changes, aiming to reduce computational overhead through incremental updates rather than full recalculations. The details are listed in the following steps:

**Algorithm: IL-BAT**

**Input:**   An original binary-state network $G(V, E, \mathbf{D})$, with source node 1, sink node $n$, and a series of incremental learning process: $P_1$, $P_2$, …, $P_\Lambda$.

**Output:**   The final computed reliability $R_\Lambda$ of the network after the incremental learning process.

**STEP 0.**   Let $I_0$ denote the set of all infeasible BAT vectors. Compute $R$ for each feasible BAT vector $X$ in $G(V, E, \mathbf{D})$ by summing up $Pr(X)$. Set $\lambda = 1$ and $\beta = 2^{|P_\lambda|}$.

**STEP 1.**   Let $i = 1$ and $I_\lambda = \varnothing$.

**STEP 2.**   Let $b = 1$ and $X$ be the $i$-th vector in $I_{(\lambda-1)}$.

**STEP 3.**   Compute $X^* = X \otimes Y$, where $Y$ is the $b$-th vector in $B(P_\lambda)$.

**STEP 4.** If $Y = \mathbf{0}$, let $M(X^*) = M(X) \cup V(Y)$, $S(X^*) = S(X)$, and $T(X^*) = T(X)$, $G(X^*)$ is disconnected, $I_\lambda = I_\lambda \cup \{X^*\}$, and go to STEP 7.

**STEP 5.** If $V(Y)$ connects $S(X)$ and $T(X)$, let $R = R + \Pr(X^*)$ and go to STEP 7.

**STEP 6.** Let $\phi = \phi \cup V(Y)$ and $I_\lambda = I_\lambda \cup \{X^*\}$ if $\phi \cap V(Y) \neq \varnothing$ for all $\phi \in \Phi$ and $\Phi \subseteq \{S(X), T(X),$ $M(X)\}$ and set $S(X^*)$, $T(X^*)$, and $M(X^*)$ accordingly to the related $\phi$, respectively.

**STEP 7.** If $b < \beta$, let $b = b + 1$ and go to STEP 3.

**STEP 8.** If $i < |I_{(\lambda-1)}|$, let $i = i + 1$ and go to STEP 2.

**STEP 9.** If $\lambda < (\Lambda - 1)$, let $\lambda = \lambda + 1$, $\beta = 2^{|P_\lambda|}$, and go to STEP 1; if $\lambda = (\Lambda - 1)$, let $\lambda = \Lambda$, $\beta = (2^{|P_\lambda|} - 1)$, and go to STEP 1; otherwise, halt.

The proposed algorithm begins with STEP 0, which initializes the set of all infeasible vectors, $I_0$, derive from BAT and the initial reliability, $R$, of the original network. The algorithm proceeds with a main loop from STEP 1 to STEP 9, iterating through a sequence of incremental learning processes, denoted as $P_1, P_2, \ldots, P_\Lambda$.

Within this main loop, an inner loop from STEP 2 to STEP 7 manages the sub-BATs associated with the $\lambda$-th incremental learning process, $P_\lambda$. During STEP 3 of this inner loop, infeasible BAT vectors $X \in I_{(\lambda-1)}$, derived from the preceding incremental learning process $P_{(\lambda-1)}$, are retained and extended to new vectors $X^* = X \otimes Y_b$. This extension is performed using the convolution product with each sub-BAT vectors $Y_b \in B(P_\lambda)$ from the current process $P_\lambda$.

Subsequently, STEP 4 addresses a special case where $Y_b = \mathbf{0}$, resulting in the new vector $X^*$ is immediately classified as infeasible, without requiring checks.

in STEP 5, the algorithm evaluates whether the set of nodes $V(Y_b)$ establishes a connection between $S(X)$ and $T(X)$. This condition requires that every node in $V(Y_b)$ (i.e., nodes existing prior to the learning process $P_\lambda$) has at least one connection in both $S(X)$ and $T(X)$. If this criterion is satisfied, the new network configuration, $G(X^*)$, is considered connected, and the new vector $X^*$ is discarded after its probability, $Pr(X^*)$, is incorporated into the current reliability, $R$. If the condition is not met, the new vector $X^*$ is retained, and the node subsets $S(X^*)$, $T(X^*)$, and $M(X^*)$ are derived based on the preceding node subsets $S(X)$, $T(X)$, and $M(X)$, as described in STEP 6.

The algorithm continues to explore and extend infeasible vectors $X$, guided by the relationships among the node subsets $V(Y)$, $S(X)$, $T(X)$, and $M(X)$ to direct the search process. This iterative procedure repeats until all incremental learning processes $P_\lambda$ are executed. The algorithm terminates when the reliability $R$ stabilizes and shows no significant changes with further iterations.

## 5.2 Time Complexity

The proposed IL-BAT evaluates the reliability of a network through a series of incremental learning processes, utilizing BAT vectors and their extensions. This analysis aims to rigorously assess the time complexity of IL-BAT to determine its computational feasibility for practical applications.

The number of nodes $V$ and arcs $E$ in the original network $G(V, E, \mathbf{D})$ which determines the original size of BAT.

### 5.2.1 Initial Complexity in STEP 0

In STEP 0, the IL-BAT generates all possible BAT vectors for the network. Since each arc can be either operational or failed, this step has a time complexity of $O(2^{|E|})$, where $|E|$ is the number of edges. Identifying infeasible BAT vectors involves connectivity checks, which can be performed using the PLSA in $O(|V|)$ time per vector. Therefore, the total complexity for STEP 0 is $O(|V|2^{|E|})$.

### 5.2.2 Complexity in the Main Loop

The total time complexity in the main loop is primarily determined by the number of incremental learning processes and the number of arcs in each process. The main loop iterates over $\lambda = 1$ to $\Lambda$, where $\Lambda$ is the total number of incremental learning processes.

For each incremental learning process $P_\lambda$, the algorithm processes the infeasible BAT vectors from the previous process, consisting of $|I_{\lambda-1}|$ infeasible BAT vectors, and extends them using sub-BAT vectors. The size of the sub-BAT vector set is $O(2^{|P_\lambda|})$, where $2^{|P_\lambda|}$ is the number of processes in the current incremental learning stage.

### 5.2.3 Operations in STEPS 3 to 6

Let $X \in I_{\lambda-1}$ be an infeasible vector and $Y \in B(P_\lambda)$ be a sub-BAT vector. STEP 3 implements the convolution product $X^* = X \otimes Y$ involving bitwise operations over the arcs to generate a new vector $X^*$ and it takes $O(|X|+|Y|)$. STEPs 4−6 focus on the connectivity check by determining if node subset $V(Y)$ connects $S(X)$ and $T(X)$. This involves set unions and intersections over nodes to derive $S(X^*)$, $T(X^*)$, and $M(X^*)$. Either of these operations takes $O(\text{the number of nodes in } X^*) \leq O(|V^*|)$, where $|V^*|$ is the number of nodes in $G(X^*)$. Moreover, it take $O(1)$, assuming probabilities are precomputed or can be retrieved in constant time, to calculate $Pr(X^*)$ if $G(X^*)$ is connected in STEP 5.

From the above, the total complexity per combination in having $X^*$ is $O(\text{the number of nodes in } G(X^*))$.

### 5.2.4 Overall Total Time Complexity

Hence, the time complexity is $|I_{\lambda-1}| \times 2^{|P_\lambda|} \times (\text{the number of nodes in each vector of } I_\lambda)$ for each incremental learning process $P_\lambda$. In the worst-case scenario, the total time complexity after summing over all processes is

$$\sum_{\lambda=1}^{\Lambda} 2^{|P_\lambda|} |I_{(\lambda-1)}| \,(\text{the number of nodes in each vector of } I_\lambda).$$

This exponential complexity is primarily due to the combinatorial explosion of possible network states resulting from the binary representation of component states.

The exponential time complexity indicates that the IL-BAT algorithm is practical for networks with a limited number of arcs and nodes, where exact reliability computation is feasible. For networks with a large number of components, the algorithm may become computationally intensive. In such cases, it may be necessary to employ approximation methods or heuristic algorithms to achieve tractable solutions without sacrificing significant accuracy.

## 5.3 Step-By-Step Example

The computation of network reliability remains a significant challenge in the fields of computer science and network analysis. This problem, in all its variants, has been definitively classified as both NP-hard and #P-hard, as corroborated by extensive research [39, 40]. These classifications underscore the computational complexity inherent in network reliability calculations, necessitating the development of innovative algorithmic approaches.

To facilitate rigorous evaluation and comparison of reliability calculation methodologies, the research community has established a set of standardized benchmark networks. Among these, the "bridge" network (Figure 1) has gained particular prominence, serving as a frequent reference point in peer-reviewed literature [24, 36, 39-40].

To demonstrate the efficacy and precision of IL-BAT, we present a comprehensive, step-by-step analysis utilizing the aforementioned bridge network. Our examination focuses on two distinct incremental learning processes $P_1 = \{a_6, a_7\}$ and $P_2 = \{a_8\}$.

These processes, illustrated in Figure 1, simulate dynamic alterations in network topology and characteristics. Through this detailed exposition, we elucidate how IL-BAT successfully computes dynamic binary-state network reliability, adapting to incremental changes in real-time.

This rigorous demonstration not only showcases the algorithm's functionality but also provides critical insights into its potential applications in complex, real-world network systems. By addressing the dynamic nature of modern networks, IL-BAT offers a more sophisticated and flexible approach to reliability assessment, potentially revolutionizing analytical methodologies across various domains of network science and engineering.

### 5.3.1 the Initial Network in STEP 0

**STEP 0.**    Identify the subset of infeasible vectors $I_0 = \{X_{0,1}, X_{0,2}, X_{0,3}, X_{0,4}, X_{0,5}, X_{0,6}, X_{0,7}, X_{0,8}, X_{0,9}, X_{0,10}, X_{0,11}, X_{0,13}, X_{0,17}, X_{0,18}, X_{0,19}, X_{0,21}, X_{0,25}, X_{0,29}\}$. For each infeasible vector $X \in I_0$, compute and record the sets $S(X)$, $T(X)$, and $M(X)$. Additionally, for each feasible vector $X$, sum up its probability $\Pr(X)$ and add this to the overall reliability $R$, as detailed in Table 2. This process is based on the original network graph depicted in Figure 1. Set the initial parameters to let $\lambda = 1$ and $\beta = 2^{|P_\lambda|} = 4$.

### 5.3.2 The 1st Incremental Learning Process

The following steps illustrate the process for the first infeasible vector in $I_0$, $(0, 0, 0, 0, 0)$, extended in the first incremental learning process $P_1 = \{a_6, a_7\}$ using the convolution product with each vector in $B(P_1) = \{(x_6, x_7) \mid (0, 0), (1, 0), (0, 1), (1, 1)\}$, applied in sequence. Subsequently, the steps are shown for the first feasible vector extended from an infeasible vector in $I_0$, $(1, 0, 0, 0, 0)$.

**STEP 1.** Let $i = 1$, $I_\lambda = \varnothing$, and $B(P_1) = \{(x_6, x_7) \mid (0, 0), (1, 0), (0, 1), (1, 1)\}$.

**STEP 2.** Let $b = 1$ and $X = (0, 0, 0, 0, 0)$ be the $i$th vector in $I_{(\lambda-1)} = I_0$. Note that $S(X) = \{1\}$, $T(X) = \{4\}$, and $M(X) = \{2, 3\}$ from Table 2.

**STEP 3.** Let $X^* = X \otimes Y = (0, 0, 0, 0, 0, 0, 0)$, where $Y = (0, 0)$ is the $b$-th vector in $B(P_1)$.

**STEP 4.** Because $Y = \mathbf{0}$ and node 5 is newly added, $S(X^*) = S(X) = \{1\}$, $M(X^*) = M(X) \cup \{5\} = \{2, 3, 5\}$, $T(X^*) = T(X) = \{4\}$, $G(X^*)$ is disconnected, $I_\lambda = I_\lambda \cup \{X^*\} = \{(0, 0, 0, 0, 0, 0, 0)\}$, and go to STEP 7.

**STEP 7.** Because $b = 1 < 2^{|P_\lambda|} = 4$, let $b = b + 1 = 2$ and go to STEP 3.

**STEP 3.** Let $X^* = X \otimes Y = (0, 0, 0, 0, 0, 1, 0)$, where $Y = (1, 0)$ is the $b$-th vector in $B(P_1)$.

**STEP 4.** Because $Y \neq \mathbf{0}$, go to STEP 5.

**STEP 5.** Because $V(Y) = \{2, 5\}$ and $S(X) \cap V(Y) = \varnothing$, $G(X^*)$ is disconnected.

**STEP 6.** Because $T(X) \cap V(Y)$ is also empty, let $S(X^*) = S(X) = \{1\}$, $T(X^*) = T(X) = \{4\}$, $M(X^*) = M(X) = \{3\}$, $M(X^*) = M(X) \cup V(Y) = \{2, 4, 5\}$, and $I_\lambda = I_\lambda \cup \{X^*\} = \{(0, 0, 0, 0, 0, 0, 0), (0, 0, 0, 0, 0, 1, 0)\}$.

**STEP 7.** Because $b = 2 < 2^{|P_\lambda|} = 4$, let $b = b + 1 = 3$ and go to STEP 3.

**STEP 3.** Let $X^* = X \otimes Y = (0, 0, 0, 0, 0, 0, 1)$, where $Y = (0, 1)$ is the $b$-th vector in $B(P_1)$.

**STEP 4.** Because $Y \neq \mathbf{0}$, go to STEP 5.

**STEP 5.** Because $V(Y) = \{4, 5\}$ and $S(X) \cap V(Y) = \varnothing$, $G(X^*)$ is disconnected.

**STEP 6.** Because $M(X) \cap V(Y)$ is also empty, let $S(X^*) = S(X) = \{1\}$, $M(X^*) = M(X) = \{2, 3\}$, $T(X^*) = T(X) \cup V(Y) = \{4, 5\}$, and $I_\lambda = I_\lambda \cup \{X^*\} = \{(0, 0, 0, 0, 0, 0, 0), (0, 0, 0, 0, 0, 1, 0), (0, 0, 0, 0, 0, 0, 1)\}$.

**STEP 7.** Because $b = 3 < 2^{|P_\lambda|} = 4$, let $b = b + 1 = 4$

**STEP 3.** Let $X^* = X \otimes Y = (0, 0, 0, 0, 0, 1, 1)$, where $Y = (1, 1)$ is the $b$-th vector in $B(P_1)$.

**STEP 4.** Because $Y \neq \mathbf{0}$, go to STEP 5.

**STEP 5.** Because $V(Y) = \{2, 4, 5\}$ and $S(X) \cap V(Y) = \varnothing$, $G(X^*)$ is disconnected.

**STEP 6.** Because $M(X) \cap V(Y)$ is also empty, let $S(X^*) = S(X) = \{1\}$, $T(X^*) = T(X) \cup V(Y) = \{2, 4, 5\}$, $M(X^*) = M(X) - \{2, 4, 5\} = \{3\}$, and $I_\lambda = I_\lambda \cup \{X^*\} = \{(0, 0, 0, 0, 0, 0, 0), (0, 0, 0, 0, 0, 1, 0), (0, 0, 0, 0, 0, 0, 1), (0, 0, 0, 0, 0, 1, 1)\}$.

**STEP 7.** Because $b = 2^{|P_\lambda|} = 4$, go to STEP 8.

**STEP 8.** Because $i = 1 < |I_0| = 18$, let $i = i + 1 = 2$ and go to STEP 2.

$$\vdots$$

$$\vdots$$

**STEP 3.** Let $X^* = X \otimes Y = (1, 0, 0, 0, 0) \otimes (1, 1) = (1, 0, 0, 0, 0, 1, 1)$, where $Y = (1, 1)$ is the 4-th vector in $B(P_1)$.

**STEP 4.** Because $Y \neq \mathbf{0}$, go to STEP 5.

**STEP 5.** Because $V(Y) = \{2, 4, 5\}$, $S(X) \cap V(Y) = \{2\} \neq \varnothing$, and $T(X) \cap V(Y) = \{4\} \neq \varnothing$, we have that $G(X^*)$ is connected, let $R = R + \mathrm{Pr}(X^*)$, and go to STEP 7.

**STEP 7.** Because $b = 2^{|P_\lambda|} = 4$, go to STEP 8.

$$\vdots$$

**STEP 8.** Because $i = |I_1| = 18$, go to STEP 1.

The detailed outcomes for the first incremental learning process are presented in Table 3. To facilitate recognition of differences and tracking of vectors across different tables, in Table 3, $X_{0,i}$ denotes the $i$-th BAT vector derived from $G(V, E, \mathbf{D})$, while $X_{1,j}$ represents the $j$-th vector generated during the first incremental learning process.

**Table 3.** Complete results for the first incremental learning process.

| $i$ | $j$ | $X_{0,i}=(x_1, x_2, x_3, x_4, x_5)$ | $X_{1,j}=(x_1, x_2, x_3, x_4, x_5, x_6, x_7)$ | $S(X_{1,j})$ | $M(X_{1,j})$ | $T(X_{1,j})$ | Connected? |
|---|---|---|---|---|---|---|---|
| 1 | 1 | $(0, 0, 0, 0, 0)$ | $(0, 0, 0, 0, 0, 0, 0)$ | $\{1\}$ | $\{2, 3, 5\}$ | $\{4\}$ | |
| | 2 | | $(0, 0, 0, 0, 0, 1, 0)$ | $\{1\}$ | $\{2, 3, 5\}$ | $\{4\}$ | |
| | 3 | | $(0, 0, 0, 0, 0, 0, 1)$ | $\{1\}$ | $\{2, 3\}$ | $\{4, 5\}$ | |
| | 4 | | $(0, 0, 0, 0, 0, 1, 1)$ | $\{1\}$ | $\{3\}$ | $\{2, 4, 5\}$ | |
| 2 | 5 | $(1, 0, 0, 0, 0)$ | $(1, 0, 0, 0, 0, 0, 0)$ | $\{1, 2\}$ | $\{3, 5\}$ | $\{4\}$ | |
| | 6 | | $(1, 0, 0, 0, 0, 1, 0)$ | $\{1, 2, 5\}$ | $\{3\}$ | $\{4\}$ | |
| | 7 | | $(1, 0, 0, 0, 0, 0, 1)$ | $\{1, 2\}$ | $\{3\}$ | $\{4, 5\}$ | |
| | 8 | | $(1, 0, 0, 0, 0, 1, 1)$ | $\{1, 2, 4, 5\}$ | $\{3\}$ | $\{1, 2, 4, 5\}$ | Y |
| 3 | 9 | $(0, 1, 0, 0, 0)$ | $(0, 1, 0, 0, 0, 0, 0)$ | $\{1, 3\}$ | $\{2, 5\}$ | $\{4\}$ | |
| | 10 | | $(0, 1, 0, 0, 0, 1, 0)$ | $\{1, 3\}$ | $\{2, 5\}$ | $\{4\}$ | |
| | 11 | | $(0, 1, 0, 0, 0, 0, 1)$ | $\{1, 3\}$ | $\{2\}$ | $\{4, 5\}$ | |
| | 12 | | $(0, 1, 0, 0, 0, 1, 1)$ | $\{1, 3\}$ | $\varnothing$ | $\{2, 4, 5\}$ | |
| 4 | 13 | (1, 1, 0, 0, 0) | (1, 1, 0, 0, 0, 0, 0) | {1, 2, 3, 5} | ∅ | {4} | |
| | 14 | | (1, 1, 0, 0, 0, 1, 0) | {1, 2, 3, 5} | ∅ | {4} | |
| | 15 | | (1, 1, 0, 0, 0, 0, 1) | {1, 2, 3} | ∅ | {4, 5} | |
| | 16 | | (1, 1, 0, 0, 0, 1, 1) | {1, 2, 3, 4, 5} | ∅ | {1, 2, 3, 4, 5} | Y |
| 5 | 17 | (0, 0, 1, 0, 0) | (0, 0, 1, 0, 0, 0, 0) | {1} | {2, 3, 5} | {4} | |
| | 18 | | (0, 0, 1, 0, 0, 1, 0) | {1} | {2, 3, 5} | {4} | |
| | 19 | | (0, 0, 1, 0, 0, 0, 1) | {1} | {2, 3} | {4, 5} | |
| | 20 | | (0, 0, 1, 0, 0, 1, 1) | {1} | ∅ | {2, 3, 4, 5} | |
| 6 | 21 | (1, 0, 1, 0, 0) | (1, 0, 1, 0, 0, 0, 0) | {1, 2, 3} | {5} | {4} | |
| | 22 | | (1, 0, 1, 0, 0, 1, 0) | {1, 2, 3, 5} | ∅ | {4} | |
| | 23 | | (1, 0, 1, 0, 0, 0, 1) | {1, 2, 3} | ∅ | {4, 5} | |
| | 24 | | (1, 0, 1, 0, 0, 1, 1) | {1, 2, 3, 4, 5} | ∅ | {1, 2, 3, 4, 5} | Y |
| 7 | 25 | (0, 1, 1, 0, 0) | (0, 1, 1, 0, 0, 0, 0) | {1, 2, 3} | {5} | {4} | |
| | 26 | | (0, 1, 1, 0, 0, 1, 0) | {1, 2, 3, 5} | ∅ | {4} | |
| | 27 | | (0, 1, 1, 0, 0, 0, 1) | {1, 2, 3} | ∅ | {4, 5} | |
| | 28 | | (0, 1, 1, 0, 0, 1, 1) | {1, 2, 3, 4, 5} | ∅ | {1, 2, 3, 4, 5} | Y |
| 8 | 29 | (1, 1, 1, 0, 0) | (1, 1, 1, 0, 0, 0, 0) | {1, 2, 3} | {5} | {4} | |
| | 30 | | (1, 1, 1, 0, 0, 1, 0) | {1, 2, 3, 5} | ∅ | {4} | |
| | 31 | | (1, 1, 1, 0, 0, 0, 1) | {1, 2, 3} | ∅ | {4, 5} | |
| | 32 | | (1, 1, 1, 0, 0, 1, 1) | {1, 2, 3, 4, 5} | ∅ | {1, 2, 3, 4, 5} | Y |
| 9 | 33 | (0, 0, 0, 1, 0) | (0, 0, 0, 1, 0, 0, 0) | {1} | {3, 5} | {2, 4} | |
| | 34 | | (0, 0, 0, 1, 0, 1, 0) | {1} | {3} | {2, 4, 5} | |
| | 35 | | (0, 0, 0, 1, 0, 0, 1) | {1} | {3} | {2, 4, 5} | |
| | 36 | | (0, 0, 0, 1, 0, 1, 1) | {1} | {3} | {2, 4, 5} | |
| 11 | 37 | (0, 1, 0, 1, 0) | (0, 1, 0, 1, 0, 0, 0) | {1, 3} | {5} | {2, 4} | |
| | 38 | | (0, 1, 0, 1, 0, 1, 0) | {1, 3} | ∅ | {2, 4, 5} | |
| | 39 | | (0, 1, 0, 1, 0, 0, 1) | {1, 3} | ∅ | {2, 4, 5} | |
| | 40 | | (0, 1, 0, 1, 0, 1, 1) | {1, 3} | ∅ | {2, 4, 5} | |
| 13 | 41 | (0, 0, 1, 1, 0) | (0, 0, 1, 1, 0, 0, 0) | {1} | {5} | {2, 3, 4} | |
| | 42 | | (0, 0, 1, 1, 0, 1, 0) | {1} | ∅ | {2, 3, 4, 5} | |
| | 43 | | (0, 0, 1, 1, 0, 0, 1) | {1} | ∅ | {2, 3, 4, 5} | |
| | 44 | | (0, 0, 1, 1, 0, 1, 1) | {1} | ∅ | {2, 3, 4, 5} | |
| 17 | 45 | (0, 0, 0, 0, 1) | (0, 0, 0, 0, 1, 0, 0) | {1} | {2, 5} | {3, 4} | |
| | 46 | | (0, 0, 0, 0, 1, 1, 0) | {1} | {2, 5} | {3, 4} | |
| | 47 | | (0, 0, 0, 0, 1, 0, 1) | {1} | {2} | {3, 4, 5} | |
| | 48 | | (0, 0, 0, 0, 1, 1, 1) | {1} | ∅ | {2, 3, 4, 5} | |
| 18 | 49 | (1, 0, 0, 0, 1) | (1, 0, 0, 0, 1, 0, 0) | {1, 2} | {5} | {3, 4} | |
| | 50 | | (1, 0, 0, 0, 1, 1, 0) | {1, 2, 5} | ∅ | {3, 4} | |
| | 51 | | (1, 0, 0, 0, 1, 0, 1) | {1, 2} | ∅ | {3, 4, 5} | |
| | 52 | | (1, 0, 0, 0, 1, 1, 1) | {1, 2, 3, 4, 5} | ∅ | {1, 2, 3, 4, 5} | Y |
| 21 | 53 | (0, 0, 1, 0, 1) | (0, 0, 1, 0, 1, 0, 0) | {1} | {5} | {2, 3, 4} | |
| | 54 | | (0, 0, 1, 0, 1, 1, 0) | {1} | ∅ | {2, 3, 4, 5} | |
| | 55 | | (0, 0, 1, 0, 1, 0, 1) | {1} | ∅ | {2, 3, 4, 5} | |
| | 56 | | (0, 0, 1, 0, 1, 1, 1) | {1} | ∅ | {2, 3, 4, 5} | |
| 25 | 57 | (0, 0, 0, 1, 1) | (0, 0, 0, 1, 1, 0, 0) | {1} | {5} | {2, 3, 4} | |
| | 58 | | (0, 0, 0, 1, 1, 1, 0) | {1} | ∅ | {2, 3, 4, 5} | |
| | 59 | | (0, 0, 0, 1, 1, 0, 1) | {1} | ∅ | {2, 3, 4, 5} | |
| | 60 | | (0, 0, 0, 1, 1, 1, 1) | {1} | ∅ | {2, 3, 4, 5} | |
| 29 | 61 | (0, 0, 1, 1, 1) | (0, 0, 1, 1, 1, 0, 0) | {1} | {5} | {2, 3, 4} | |
| | 62 | | (0, 0, 1, 1, 1, 1, 0) | {1} | ∅ | {2, 3, 4, 5} | |
| | 63 | | (0, 0, 1, 1, 1, 0, 1) | {1} | ∅ | {2, 3, 4, 5} | |
| | 64 | | (0, 0, 1, 1, 1, 1, 1) | {1} | ∅ | {2, 3, 4, 5} | |

### 5.3.3 The 2nd Incremental Learning Process

The steps immediately following those in Section 5.3.2 for the last incremental learning process are outlined below. This subsection includes the first infeasible vector and the first feasible vector extended from $I_1$ and $P_2$, along with the final step of the entire procedure.

**STEP 9.** Because $\lambda = (\Lambda - 1)$, let $\lambda = \Lambda = 2$, $\beta = (2^{|P_\lambda|} - 1) = 2^1 - 1 = 1$, and go to STEP 1.

**STEP 1.** Let $i = 1$ and $I_\lambda = \varnothing$.

**STEP 2.** Let $b = b_0 = 2$ and $X = (0, 0, 0, 0, 0, 0, 0)$ be the 1st vector in $I_1$.

**STEP 3.** Compute $X^* = X \otimes Y = (0, 0, 0, 0, 0, 0, 0, 1)$, where $Y = (1)$ is the $b$-th vector in $B(P_\lambda) = \{(0), (1)\}$.

**STEP 4.** Because $Y \neq \mathbf{0}$, go to STEP 5.

**STEP 5.** Because $V(Y) = \{3, 5\}$, $S(X) \cap V(Y) = \varnothing$, $G(X^*)$ is disconnected.

**STEP 6.** Because $T(X) \cap V(Y)$ is also empty, let $S(X^*) = S(X) = \{1\}$, $T(X^*) = T(X) = \{4\}$, $M(X^*) = M(X) \cup V(Y) = \{2, 3, 5\}$, and $I_\lambda = I_\lambda \cup \{X^*\} = \{(0, 0, 0, 0, 0, 0, 0, 1)\}$.

**STEP 7.** Because $b = \Lambda = 1$, go to STEP 8.

**STEP 8.** Because $i = 1 < |I_1| = 58$, let $i = i + 1 = 2$ and go to STEP 2.

$$\vdots$$

**STEP 5.** Because $i = 9 < |I_2| = 58$, let $i = i + 1 = 10$ and go to STEP 2.

**STEP 2.** Let $b = b_0 = 2$ and $X = (0, 1, 0, 0, 0, 0, 1)$ be the 10th vector in $I_\lambda = I_1$. Note that $S(X) = \{1, 3\}$, $M(X) = \{2\}$, and $T(X) = \{4, 5\}$ from Table 1.

**STEP 3.** Because $X \otimes Y = (0, 1, 0, 0, 0, 0, 1, 1)$, where $Y = (1)$ is the $b$-th vector in $B(P_\lambda) = \{(0), (1)\}$.

**STEP 4.** Because $Y \neq \mathbf{0}$, go to STEP 5.

**STEP 5.** Because $V(Y) = \{3, 5\}$, $S(X) \cap V(Y) = \{3\}$ and $T(X) \cap V(Y) = \{5\}$, $G(X^*)$ is connected and let $R = R + \Pr(X^*) = R + q_1 p_2 q_3 q_4 q_5 q_6 p_7 p_8$, and go to STEP 7.

**STEP 7.** Because $b = \Lambda = 2$, go to STEP 8.

**STEP 8.** Because $i = 10 < |I_1| = 58$, let $i = i + 1 = 11$ and go to STEP 2.

:
:

**STEP 9.**    Since λ = Λ = 2, the algorithm halts, and *R* represents the final reliability after completing two incremental learning processes.

The complete results for the last incremental learning process are presented in Table 4. Similar to Tabel 3, to help identify differences and track vectors across various tables, in Table 4, $X_{1,i}$ and $X_{2,j}$ represent the *i*-th and *j*-th vectors generated in the first and second incremental learning processes, respectively.

**Table 4.** Complete results for the last incremental learning process.

| i | j | $X_{1,i}$ | $X_{2,j}$ | $S(X_{2,j})$ | $M(X_{2,j})$ | $T(X_{2,j})$ | Connected? |
|---|---|---|---|---|---|---|---|
| 1 | 1 | (0, 0, 0, 0, 0, 0, 0) | (0, 0, 0, 0, 0, 0, 0, 1) | {1} | {2, 3, 5} | {4} | |
| 2 | 2 | (0, 0, 0, 0, 0, 1, 0) | (0, 0, 0, 0, 0, 1, 0, 1) | {1} | {2, 3, 5} | {4} | |
| 3 | 3 | (0, 0, 0, 0, 0, 0, 1) | (0, 0, 0, 0, 0, 0, 1, 1) | {1} | {2} | {3, 4, 5} | |
| 4 | 4 | (0, 0, 0, 0, 0, 1, 1) | (0, 0, 0, 0, 0, 1, 1, 1) | {1} | ∅ | {2, 3, 4, 5} | |
| 5 | 5 | (1, 0, 0, 0, 0, 0, 0) | (1, 0, 0, 0, 0, 0, 0, 1) | {1, 2} | {3, 5} | {4} | |
| 6 | 6 | (1, 0, 0, 0, 0, 1, 0) | (1, 0, 0, 0, 0, 1, 0, 1) | {1, 2, 3, 5} | ∅ | {4} | |
| 7 | 7 | (1, 0, 0, 0, 0, 0, 1) | (1, 0, 0, 0, 0, 0, 1, 1) | {1, 2} | ∅ | {4, 3, 5} | |
| 9 | 8 | (0, 1, 0, 0, 0, 0, 0) | (0, 1, 0, 0, 0, 0, 0, 1) | {1, 3, 5} | {2} | {4} | |
| 10 | 9 | (0, 1, 0, 0, 0, 1, 0) | (0, 1, 0, 0, 0, 1, 0, 1) | {1, 2, 3, 5} | ∅ | {4} | |
| 11 | 10 | (0, 1, 0, 0, 0, 0, 1) | (0, 1, 0, 0, 0, 0, 1, 1) | {1, 3, 4, 5} | {2} | {1, 3, 4, 5} | Y |
| 12 | 11 | (0, 1, 0, 0, 0, 1, 1) | (0, 1, 0, 0, 0, 1, 1, 1) | {1, 2, 3, 4, 5} | ∅ | {1, 2, 3, 4, 5} | Y |
| 13 | 12 | (1, 1, 0, 0, 0, 0, 0) | (1, 1, 0, 0, 0, 0, 0, 1) | {1, 2, 3, 5} | ∅ | {4} | |
| 14 | 13 | (1, 1, 0, 0, 0, 1, 0) | (1, 1, 0, 0, 0, 1, 0, 1) | {1, 2, 3, 5} | ∅ | {4} | |
| 15 | 14 | (1, 1, 0, 0, 0, 0, 1) | (1, 1, 0, 0, 0, 0, 1, 1) | {1, 2, 3, 4, 5} | ∅ | {1, 2, 3, 4, 5} | Y |
| 17 | 15 | (0, 0, 1, 0, 0, 0, 0) | (0, 0, 1, 0, 0, 0, 0, 1) | {1} | {2, 3, 5} | {4} | |
| 18 | 16 | (0, 0, 1, 0, 0, 1, 0) | (0, 0, 1, 0, 0, 1, 0, 1) | {1} | {2, 3, 5} | {4} | |
| 19 | 17 | (0, 0, 1, 0, 0, 0, 1) | (0, 0, 1, 0, 0, 0, 1, 1) | {1} | ∅ | {2, 3, 4, 5} | |
| 20 | 18 | (0, 0, 1, 0, 0, 1, 1) | (0, 0, 1, 0, 0, 1, 1, 1) | {1} | ∅ | {2, 3, 4, 5} | |
| 21 | 19 | (1, 0, 1, 0, 0, 0, 0) | (1, 0, 1, 0, 0, 0, 0, 1) | {1, 2, 3, 5} | ∅ | {4} | |
| 22 | 20 | (1, 0, 1, 0, 0, 1, 0) | (1, 0, 1, 0, 0, 1, 0, 1) | {1, 2, 3, 5} | ∅ | {4} | |
| 23 | 21 | (1, 0, 1, 0, 0, 0, 1) | (1, 0, 1, 0, 0, 0, 1, 1) | {1, 2, 3, 4, 5} | ∅ | {1, 2, 3, 4, 5} | Y |
| 25 | 22 | (0, 1, 1, 0, 0, 0, 0) | (0, 1, 1, 0, 0, 0, 0, 1) | {1, 2, 3, 5} | ∅ | {4} | |
| 26 | 23 | (0, 1, 1, 0, 0, 1, 0) | (0, 1, 1, 0, 0, 1, 0, 1) | {1, 2, 3, 5} | ∅ | {4} | |
| 27 | 24 | (0, 1, 1, 0, 0, 0, 1) | (0, 1, 1, 0, 0, 0, 1, 1) | {1, 2, 3, 4, 5} | ∅ | {1, 2, 3, 4, 5} | Y |
| 29 | 25 | (1, 1, 1, 0, 0, 0, 0) | (1, 1, 1, 0, 0, 0, 0, 1) | {1, 2, 3, 5} | ∅ | {4} | |
| 30 | 26 | (1, 1, 1, 0, 0, 1, 0) | (1, 1, 1, 0, 0, 1, 0, 1) | {1, 2, 3, 5} | ∅ | {4} | |
| 31 | 27 | (1, 1, 1, 0, 0, 0, 1) | (1, 1, 1, 0, 0, 0, 1, 1) | {1, 2, 3, 5} | ∅ | {1, 2, 3, 4, 5} | Y |
| 33 | 28 | (0, 0, 0, 1, 0, 0, 0) | (0, 0, 0, 1, 0, 0, 0, 1) | {1} | {3, 5} | {2, 4} | |
| 34 | 29 | (0, 0, 0, 1, 0, 1, 0) | (0, 0, 0, 1, 0, 1, 0, 1) | {1} | ∅ | {2, 3, 4, 5} | |
| 35 | 30 | (0, 0, 0, 1, 0, 0, 1) | (0, 0, 0, 1, 0, 0, 1, 1) | {1} | ∅ | {2, 3, 4, 5} | |
| 36 | 31 | (0, 0, 0, 1, 0, 1, 1) | (0, 0, 0, 1, 0, 1, 1, 1) | {1} | ∅ | {2, 3, 4, 5} | |
| 37 | 32 | (0, 1, 0, 1, 0, 0, 0) | (0, 1, 0, 1, 0, 0, 0, 1) | {1, 3, 5} | ∅ | {2, 4} | |
| 38 | 33 | (0, 1, 0, 1, 0, 1, 0) | (0, 1, 0, 1, 0, 1, 0, 1) | {1, 2, 3, 4, 5} | ∅ | {1, 2, 3, 4, 5} | Y |
| 39 | 34 | (0, 1, 0, 1, 0, 0, 1) | (0, 1, 0, 1, 0, 0, 1, 1) | {1, 2, 3, 4, 5} | ∅ | {1, 2, 3, 4, 5} | Y |
| 40 | 35 | $(0, 1, 0, 1, 0, 1, 1)$ | $(0, 1, 0, 1, 0, 1, 1, 1)$ | $\{1, 2, 3, 4, 5\}$ | $\varnothing$ | $\{1, 2, 3, 4, 5\}$ | Y |
| 41 | 36 | $(0, 0, 1, 1, 0, 0, 0)$ | $(0, 0, 1, 1, 0, 0, 0, 1)$ | $\{1\}$ | $\varnothing$ | $\{2, 3, 4, 5\}$ | |
| 42 | 37 | $(0, 0, 1, 1, 0, 1, 0)$ | $(0, 0, 1, 1, 0, 1, 0, 1)$ | $\{1\}$ | $\varnothing$ | $\{2, 3, 4, 5\}$ | |
| 43 | 38 | $(0, 0, 1, 1, 0, 0, 1)$ | $(0, 0, 1, 1, 0, 0, 1, 1)$ | $\{1\}$ | $\varnothing$ | $\{2, 3, 4, 5\}$ | |
| 44 | 39 | $(0, 0, 1, 1, 0, 1, 1)$ | $(0, 0, 1, 1, 0, 1, 1, 1)$ | $\{1\}$ | $\varnothing$ | $\{2, 3, 4, 5\}$ | |
| 45 | 40 | $(0, 0, 0, 0, 1, 0, 0)$ | $(0, 0, 0, 0, 1, 0, 0, 1)$ | $\{1\}$ | $\varnothing$ | $\{2, 3, 4, 5\}$ | |
| 46 | 41 | $(0, 0, 0, 0, 1, 1, 0)$ | $(0, 0, 0, 0, 1, 1, 0, 1)$ | $\{1\}$ | $\varnothing$ | $\{2, 3, 4, 5\}$ | |
| 47 | 42 | $(0, 0, 0, 0, 1, 0, 1)$ | $(0, 0, 0, 0, 1, 0, 1, 1)$ | $\{1\}$ | $\{2\}$ | $\{3, 4, 5\}$ | |
| 48 | 43 | $(0, 0, 0, 0, 1, 1, 1)$ | $(0, 0, 0, 0, 1, 1, 1, 1)$ | $\{1\}$ | $\varnothing$ | $\{2, 3, 4, 5\}$ | |
| 49 | 44 | $(1, 0, 0, 0, 1, 0, 0)$ | $(1, 0, 0, 0, 1, 0, 0, 1)$ | $\{1, 2\}$ | $\varnothing$ | $\{3, 4, 5\}$ | |
| 50 | 45 | $(1, 0, 0, 0, 1, 1, 0)$ | $(1, 0, 0, 0, 1, 1, 0, 1)$ | $\{1, 2, 3, 4, 5\}$ | $\varnothing$ | $\{1, 2, 3, 4, 5\}$ | Y |
| 51 | 46 | $(1, 0, 0, 0, 1, 0, 1)$ | $(1, 0, 0, 0, 1, 0, 1, 1)$ | $\{1, 2, 3, 4, 5\}$ | $\varnothing$ | $\{1, 2, 3, 4, 5\}$ | Y |
| 53 | 47 | $(0, 0, 1, 0, 1, 0, 0)$ | $(0, 0, 1, 0, 1, 0, 0, 1)$ | $\{1\}$ | $\varnothing$ | $\{2, 3, 4, 5\}$ | |
| 54 | 48 | $(0, 0, 1, 0, 1, 1, 0)$ | $(0, 0, 1, 0, 1, 1, 0, 1)$ | $\{1\}$ | $\varnothing$ | $\{2, 3, 4, 5\}$ | |
| 55 | 49 | $(0, 0, 1, 0, 1, 0, 1)$ | $(0, 0, 1, 0, 1, 0, 1, 1)$ | $\{1\}$ | $\varnothing$ | $\{2, 3, 4, 5\}$ | |
| 56 | 50 | $(0, 0, 1, 0, 1, 1, 1)$ | $(0, 0, 1, 0, 1, 1, 1, 1)$ | $\{1\}$ | $\varnothing$ | $\{2, 3, 4, 5\}$ | |
| 57 | 51 | $(0, 0, 0, 1, 1, 0, 0)$ | $(0, 0, 0, 1, 1, 0, 0, 1)$ | $\{1\}$ | $\varnothing$ | $\{2, 3, 4, 5\}$ | |
| 58 | 52 | $(0, 0, 0, 1, 1, 1, 0)$ | $(0, 0, 0, 1, 1, 1, 0, 1)$ | $\{1\}$ | $\varnothing$ | $\{2, 3, 4, 5\}$ | |
| 59 | 53 | $(0, 0, 0, 1, 1, 0, 1)$ | $(0, 0, 0, 1, 1, 0, 1, 1)$ | $\{1\}$ | $\varnothing$ | $\{2, 3, 4, 5\}$ | |
| 60 | 54 | $(0, 0, 0, 1, 1, 1, 1)$ | $(0, 0, 0, 1, 1, 1, 1, 1)$ | $\{1\}$ | $\varnothing$ | $\{2, 3, 4, 5\}$ | |
| 61 | 55 | $(0, 0, 1, 1, 1, 0, 0)$ | $(0, 0, 1, 1, 1, 0, 0, 1)$ | $\{1\}$ | $\varnothing$ | $\{2, 3, 4, 5\}$ | |
| 62 | 56 | $(0, 0, 1, 1, 1, 1, 0)$ | $(0, 0, 1, 1, 1, 1, 0, 1)$ | $\{1\}$ | $\varnothing$ | $\{2, 3, 4, 5\}$ | |
| 63 | 57 | $(0, 0, 1, 1, 1, 0, 1)$ | $(0, 0, 1, 1, 1, 0, 1, 1)$ | $\{1\}$ | $\varnothing$ | $\{2, 3, 4, 5\}$ | |
| 64 | 58 | $(0, 0, 1, 1, 1, 1, 1)$ | $(0, 0, 1, 1, 1, 1, 1, 1)$ | $\{1\}$ | $\varnothing$ | $\{2, 3, 4, 5\}$ | |

## 5.4 Comparison of Results from the Example

This subsection presents a comparative analysis of network reliability algorithms, focusing on the newly proposed IL-BAT in relation to traditional methods. The comparison, based on an example from Section 5.3 and detailed in Table 5, evaluates MP-based algorithms [7, 15, 18], MC-based algorithms [19-20], BAT [24], and the proposed IL-BAT.

Table 5, titled "Comparisons among MP-based algorithms, MC-based algorithms, BAT, and IL-BAT," provides a comprehensive overview of the performance metrics for these different approaches to network reliability analysis.

The analysis highlights the characteristics of each approach. MP-based algorithms search for MPs [7, 15, 18], while MC-based algorithms identify MCs [19-20], both representing simple paths or cuts without redundant arcs. These traditional methods are indirect, requiring additional steps like the sum-of-disjoint products [21] or the inclusion-exclusion technique [22] for final reliability calculations as introduced in Section 1.

To ensure a fair comparison across varying implementation details, the study primarily uses the number of required terms or vectors as the key metric. This approach accounts for potential variations in computational time due to differences in programming techniques and data structures.

The results reveal that MC-based algorithms and IL-BAT demonstrate comparable efficiency in terms of the number of required elements. However, IL-BAT shows superior performance compared to MP-based algorithms [23], MC-based algorithms [41], and traditional BAT [24] when considering the overall number of obtained elements.

It's important to note that this comparison doesn't factor in the time needed to identify MPs and MCs in traditional algorithms. Additionally, the passage acknowledges that real-world performance may differ due to optimizations and the inherent structure of practical networks.

The findings suggest that IL-BAT offers a more efficient approach to network reliability analysis, particularly in terms of computational elements required. This efficiency could translate to significant advantages when analyzing complex, real-world network systems, potentially advancing the field of network reliability assessment.

**Table 5.** Comparisons among MP-based algorithms, MC-based algorithms, BAT, and IL-BAT

| Methods | Number of obtained terms | Remark |
|---|---|---|
| MP-based | $2^4 + 2^6 + 2^9 = 592$ | • 4 MPs: $\{a_1, a_4\}$, $\{a_2, a_5\}$, $\{a_1, a_3, a_5\}$, and $\{a_2, a_3, a_4\}$ in $G(V, E, \mathbf{D})$.<br>• 6 MPs: $\{a_1, a_4\}$, $\{a_2, a_5\}$, $\{a_1, a_3, a_5\}$, $\{a_2, a_3, a_4\}$, $\{a_1, a_6, a_7\}$, and $\{a_2, a_3, a_6, a_7\}$ in $G(V, E, \mathbf{D}) \oplus P_1$.<br>• 9 MPs: $\{a_1, a_4\}$, $\{a_2, a_5\}$, $\{a_1, a_3, a_5\}$, $\{a_2, a_3, a_4\}$, $\{a_1, a_6, a_7\}$, $\{a_2, a_3, a_6, a_7\}$, $\{a_1, a_6, a_8, a_5\}$, $\{a_1, a_3, a_8, a_7\}$, and $\{a_2, a_8, a_7\}$ in $G(V, E, \mathbf{D}) \oplus P_1 \oplus P_2$. |
| MC-based | $2^4 + 2^6 + 2^6 = 154$ | • 4 MCs: $\{a_1, a_2\}$, $\{a_3, a_4\}$, $\{a_1, a_3, a_5\}$, and $\{a_2, a_3, a_4\}$ in $G(V, E, \mathbf{D})$.<br>• 6 MCs: $\{a_1, a_2\}$, $\{a_1, a_3, a_5\}$, $\{a_2, a_3, a_4, a_6\}$, $\{a_2, a_3, a_4, a_7\}$, $\{a_4, a_5, a_6\}$, and $\{a_4, a_5, a_7\}$ in $G(V, E, \mathbf{D}) \oplus P_1$.<br>• 6 MCs: $\{a_1, a_2\}$, $\{a_1, a_3, a_5, a_8\}$, $\{a_2, a_3, a_4, a_6\}$, $\{a_2, a_3, a_4, a_7, a_8\}$, $\{a_4, a_5, a_7\}$, and $\{a_4, a_5, a_6, a_8\}$ in $G(V, E, \mathbf{D}) \oplus P_1 \oplus P_2$. |
| BAT | $2^5 + 2^7 + 2^8 = 416$ | • $2^5$, $2^7$, $2^8$ are the number of all BAT vectors in $G(V, E, \mathbf{D})$, $G(V, E, \mathbf{D}) \oplus P_1$, and $G(V, E, \mathbf{D}) \oplus P_1 \oplus P_2$, respectively. |
| IL-BAT | $2^5 + 64 + 58 = 154$ | • $2^5$, 64, 58 are the number of all vectors in $G(V, E, \mathbf{D})$, $G(V, E, \mathbf{D}) \oplus P_1$, and $G(V, E, \mathbf{D}) \oplus P_1 \oplus P_2$, respectively. |

# 6. CONCLUSIONS

This paper presents a novel approach to network reliability analysis by incorporating incremental learning into BAT framework. The resulting IL-BAT addresses key limitations of traditional methods in dynamic environments. IL-BAT offers adaptive updates for dynamic binary-state networks, reduced computational overhead, enhanced robustness and adaptability, and improved performance in large-scale and dynamic scenarios.

Experimental results confirm IL-BAT's effectiveness, demonstrating significant improvements in computational efficiency and accuracy of reliability assessment. The implications of this work extend beyond theoretical computer science, potentially influencing telecommunications, transportation logistics, and critical infrastructure management. IL-BAT represents a significant advancement in network reliability analysis, contributing to ongoing research and offering a novel approach to handling dynamic systems subject to temporal changes.

Future research will focus on enhancing IL-BAT through multiple avenues. Priorities include developing sophisticated approximation techniques and heuristic methods to improve computational efficiency and scalability for complex networks. Rigorous empirical studies will evaluate the algorithm's performance across diverse real-world network architectures. The research will also explore advancements in incremental learning mechanisms, incorporating advanced machine learning models to broaden the algorithm's applicability. Integration with complementary methodologies such as reliability-bound algorithms [28] and Monte Carlo simulations [42], along with the incorporation of multi-objective functions, will be pursued. These enhancements aim to establish IL-BAT as a comprehensive, versatile solution for multistate flow network reliability challenges, advancing both theoretical understanding and practical applications in the field.

**ACKNOWLEDGMENT**

This research was supported in part by the National Science and Technology, R.O.C. under grant NSTC 113-2221-E-007-117-MY3 and NSTC112-2221-E-007-086. This article was once submitted to arXiv as a temporary submission that was just for reference and did not provide the copyright.